\documentclass[letterpaper, 10 pt, conference]{ieeeconf}
\IEEEoverridecommandlockouts
\usepackage[utf8]{inputenc}
\usepackage[noend]{algpseudocode}
\usepackage{amsmath}
\usepackage{amssymb}    
\usepackage[utf8]{inputenc}
\usepackage{xcolor}
\usepackage{graphicx}
\usepackage{hyperref}
\usepackage{booktabs}
\usepackage{color}
\usepackage{array}
\usepackage{booktabs}
\usepackage{multirow}
\usepackage{hyperref}
\usepackage{CJKutf8}
\usepackage{amssymb} 
\usepackage{threeparttable}
\usepackage{bbding}
\usepackage[space]{cite}
\usepackage[linesnumbered,boxed,ruled,commentsnumbered]{algorithm2e}
\graphicspath{{figures/}}
\DeclareGraphicsExtensions{.pdf,.png,.jpg,.eps}

\begin{document}
\title{\Large \bf FAST-LIVO2 on Resource-Constrained Platforms: LiDAR-Inertial-Visual Odometry with Efficient Memory and Computation}

\author{
Bingyang Zhou$^{\ast 1}$, Chunran Zheng$^{\ast 1}$,  
Ziming Wang$^{1}$, Fangcheng Zhu$^{1}$, Yixi Cai$^{2}$ and Fu Zhang$^{\dagger 1}$  
\thanks{$^{\ast}$These authors contributed equally.}%
\thanks{$^{\dagger}$Corresponding author: Fu Zhang (email: \texttt{fuzhang@hku.hk}).}%
\thanks{$^{1}$Mechatronics and Robotic Systems (MaRS) Laboratory, Department of Mechanical Engineering, University of Hong Kong, Hong Kong SAR, China.}%
\thanks{$^{2}$Division of Robotics, Perception and Learning, KTH Royal Institute of Technology, Stockholm, Sweden.}%
}

	\markboth{Journal of \LaTeX\ Class Files,~Vol.~6, No.~1, January~2007}%
	{Shell \MakeLowercase{\textit{et al.}}: Bare Demo of IEEEtran.cls for Journals}
	
	\maketitle
        \pagestyle{empty}  
	\thispagestyle{empty}
	
\begin{abstract}
This paper presents a lightweight LiDAR-inertial-visual odometry system optimized for resource-constrained platforms. 
It integrates a degeneration-aware adaptive visual frame selector into error-state iterated Kalman filter (ESIKF) with sequential updates, improving computation efficiency significantly {while maintaining a similar level of robustness.} Additionally, a memory-efficient mapping structure combining a locally unified visual-LiDAR map and a long-term visual map 
{achieves a good trade-off between performance and memory usage}. Extensive experiments on x86 and ARM platforms demonstrate the system's robustness and efficiency. On the Hilti dataset, our system achieves a \textbf{33\% reduction in per-frame runtime} and \textbf{47\% lower memory usage} compared to FAST-LIVO2, with only a \textbf{3 cm increase in RMSE}. Despite this slight accuracy trade-off, our system remains competitive, outperforming state-of-the-art (SOTA) LIO methods such as FAST-LIO2 and most existing LIVO systems. These results validate the system's capability for scalable deployment on resource-constrained edge computing platforms.
\end{abstract}
	
	\IEEEpeerreviewmaketitle
	
\section{Introduction}
Odometry\cite{zhang2014loam, qin2018vins, xu2020fastlio, shan2021lvi, zheng2022fast, zheng2024fast, lin2021r2live, lin2022r,forster2016svo,geneva2020openvins} has become an essential component of robotics applications, enabling autonomous systems to simultaneously map their environments and localize themselves. It plays a pivotal role in tasks such as trajectory planning\cite{zhou2020ego} and motion control\cite{lu2022manifold}. Recent advancements in odometry algorithms have significantly improved their accuracy, with many frameworks leveraging single-modality sensors like cameras or LiDAR. However, while visual and LiDAR odometry systems excel in specific domains, each faces inherent limitations that restrict their performance in diverse scenarios, such as LiDAR struggling in structureless environments and cameras failing in textureless scenes.

To address these challenges, multi-sensor fusion approaches, particularly LiDAR-inertial-visual odometry (LIVO), have excelled in tackling the limitations of single-sensor systems. These methods integrate complementary sensors' data to enhance localization robustness across various environments.  However, most current LIVO systems\cite{shan2021lvi, lin2022r} rely on computationally intensive processes, necessitating high-performance CPUs to meet the real-time demands of sensor fusion, mapping, and optimization. The SOTA LIVO system, FAST-LIVO2\cite{zheng2024fast}, employs direct methods for state estimation leveraging all LiDAR and visual measurements within a sequential update ESIKF framework, and integrates a unified LiDAR-visual local map, achieving remarkable levels of computational efficiency and accuracy. Nevertheless, LIVO systems, including FAST-LIVO2, still require significant computational and memory resources.

As robotics applications expand to lightweight and portable platforms (e.g., drones, autonomous vehicles, and IoT devices), deploying odometry on edge computing platforms becomes increasingly critical. Edge platforms, such as ARM architectures, offer advantages like energy efficiency and widespread adoption in embedded systems, making them attractive for scalable and cost-effective solutions. However, their constrained computational power, memory, and energy efficiency impose significant challenges to traditional odometry frameworks to ensure resource-efficient performance.
\begin{figure}
    \centering
    \includegraphics[width=1.0\linewidth]{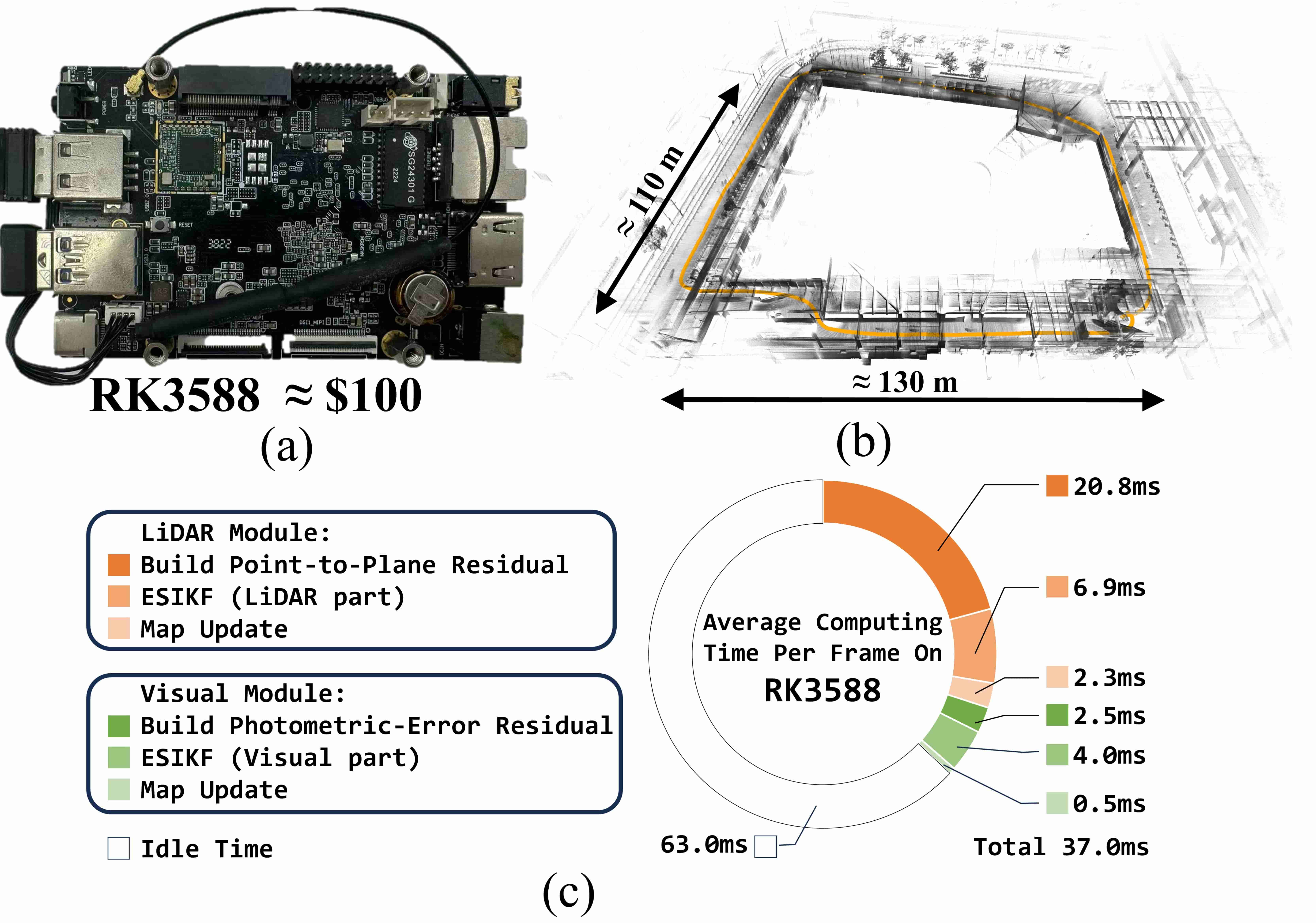}
    \caption{Overview of a real-time experiment on the low-power ARM platform. (a) The ARM platform RK3588 {with an octa-core architecture (4× Cortex-A76 + 4× Cortex-A55) and a maximum frequency of 2.4GHz} used for testing, priced at approximately 100 USD. (b) Point cloud of the nighttime street scene used for testing, with the orange trajectory representing the collected path. (c) Detailed per-frame runtime statistics of the system, with data input from both LiDAR and camera at 10 Hz (100 ms per frame).}
    \label{fig:teaser}
    \vspace{-0.8cm}
\end{figure}
Despite these challenges, edge computing brings unique benefits for real-time odometry, such as reduced latency and network dependency through localized data processing. 

{To address these advantages, we improves the state of the LIVO system, FAST-LIVO2, by minimizing its computational and memory overhead while maintaining its robustness and performances. The primary contributions of this work, when compared to FAST-LIVO2, are as follows:}
\begin{figure*}[thb]
    \centering
    \includegraphics[width=1.9\columnwidth]{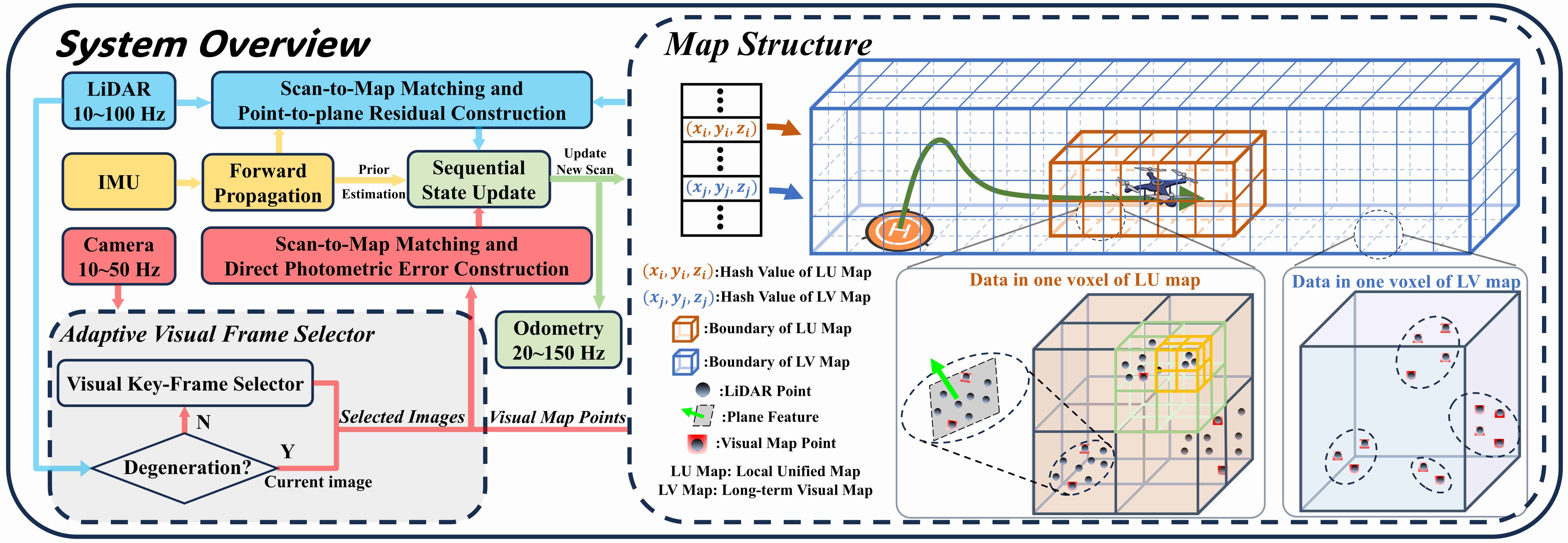}
    \caption{System overview. 
In the figure detailing voxel data on the lower-right side, the dashed ellipse encloses points considered to lie on a plane with varying scales.}
    \label{overview_framework}
    \vspace{-0.6cm}
\end{figure*}
\begin{enumerate}
    \item {\textit{LiDAR-degeneration-aware, adaptive visual update:} Rather than naturally using all available visual measurements, a LiDAR-degeneration-aware adaptive visual frame selector is integrated into the ESIKF framework of FAST-LIVO2, significantly reducing the computational burden while maintaining a similar level of odometry accuracy.}
    
    \item {\textit{Memory-efficient hybrid map structure:} we propose a hybrid map structure combines a compact local visual-LiDAR map with a long-term visual map. The latter stores sparse historical image observations to ensure odometry robustness while allowing the former to be maintained at a significantly smaller scale, reducing memory overhead. Additionally, the adaptive visual update minimizes visual features added to the map when LiDAR constraints are sufficient, further lowering the memory usage.}
    
    \item {\textit{Extensive experiments validation and dataset open-source:} Validation was performed on both public and private datasets, with the private datasets featuring more challenging and aggressive scenarios. These private datasets will be open-sourced to benefit the community. Experiments were conducted on x86 personal laptops and cost-effective ARM devices, demonstrating superior efficiency.}
    
\end{enumerate}

\section{Related works}
The topic of this work is most relevant to degeneration-aware strategy and map structure in odometry system, which are discussed as follows.

\subsection{Degeneracy-Aware Odometry Systems}
Existing studies have explored LiDAR odometry degeneracy and developed adaptive strategies within odometry systems, leveraging updated degeneration states to enhance robustness in challenging environments. Zhang et al.\cite{zhang2016degeneracy} evaluated degeneracy of optimization-based methods through analysis of geometric structure. Tuna et al.\cite{tuna2023x} proposed X-ICP, integrating localizability detection and localizability aware optimization based on the classical mechanics of a point cloud. AdaLIO\cite{lim2023adalio} adaptively adjust some key parameters after checking degeneracy. As-lio\cite{zhang2024lio}, being a variable frequency odometry, achieves great performance in some degradation scenario using a spatial overlap guided adaptive sliding window. Zhu et al.\cite{zhu2023swarm, zhu2024swarm} proposed a degeneration evaluation method to dynamically adjust the state vector to ensure robust and efficient swarm state estimation. More recently, Lee et al.\cite{lee2024switch} implemented a non-heuristic degeneracy detection using a predefined threshold on normalized eigenvalues and further use it to select a better initial guess in a switch structure. 

While previous works primarily focus on odometry robustness, our system harnesses the potential of degeneration-aware factors to dynamically adjust computational burdern allocation based on environmental constraints, achieving both robustness and significant computational efficiency.

\subsection{Map Structure in LiDAR-Visual-(Inertial) Odometry}
In LiDAR-visual-(inertial) odometry, the map structure is a critical factor for determining both efficiency and accuracy. Generally, map structures are categorized into three types. The first type maintains two separate maps for the visual and LiDAR components, where the visual backend generates 3D points independently. In the second type, two maps are still maintained, but the visual backend does not generate 3D points independently. The third type employs a unified map for both visual and LiDAR measurements.

For instance, in LVI-SAM~\cite{shan2021lvi} and R2LIVE~\cite{lin2021r2live}, VINS-mono~\cite{qin2018vins} serves as the visual subsystem, which falls under the first category. In the second category, systems like SDV-LOAM~\cite{yuan2023sdv}, DV-LOAM~\cite{wang2021dv}, FAST-LIVO~\cite{zheng2022fast}, PL-LIVO~\cite{shi2024point}, and CamVox~\cite{zhu2021camvox} reuse LiDAR points to provide depth information for visual features, patches, or line patches. However, these systems still store visual and LiDAR map points in two distinct data structures. Notably, SDV-LOAM and DV-LOAM limit the use of visual map points to keyframes or the last frame, whereas FAST-LIVO and PL-LIVO incorporate visual map points from the global map that fall within the current field of view (FoV).

In contrast, the third category includes SR-LIVO~\cite{yuan2024sr}, R3LIVE~\cite{lin2022r}, FAST-LIVO2~\cite{zheng2024fast}, and our system, which all maintain a single unified map. SR-LIVO and R3LIVE adopt an incremental k-d tree (ikd-tree\cite{cai2021ikd}) to manage point-based maps, tagging each point with RGB color. Meanwhile, FAST-LIVO2 employs a hash-indexed octree to store a surfel-based map, where each leaf node contains a LiDAR plane feature augmented with an image patch. Our system further advances this approach by maintaining a compact unified local map and preserving a lightweight long-term visual map, enabling robust odometry performance with significantly reduced memory consumption.
\section{System overview}
The system overview is shown in Fig. \ref{overview_framework}. 
At measurement level, as the input of this system, 
high-frequency raw LiDAR points are segmented into distinct scans employing scan recombination. To optimize computational efficiency, an adaptive visual frame selector dynamically selects images based on environmental constraints, allowing our degeneration-aware odometry to allocate computational resources adaptively.
For state estimation, we construct specific residuals, including LiDAR point-to-plane residuals and photometric errors, and tightly couple all sensor measurements through an ESIKF with sequential updates.
At the mapping level, a robocentric, compact unified local map and a lightweight long-term visual map are maintained through efficient integration of new observations, sliding operations, and visual features rearrangement, ensuring memory efficiency and robust performance.
\vspace{-0.15cm}
\section{METHODOLOGY\label{method}}
Previous work FAST-LIVO2\cite{zheng2024fast} established the core framework for multi-sensor (LiDAR-inertial-visual) integration, including state definitions, discrete transition models, LiDAR-visual map management, and state estimation by fusing camera and LiDAR measurements. Expanding on this foundation, this work focuses on building a lightweight LIVO system deploying on resource-constraint platforms. Therefore, in this section, we emphasize our contributions on achieving computationally efficient state estimation (Section \ref{state_est}) and memory-saving map (Section \ref{map_manager}).
\subsection{State Estimation with Visual Frame Selector\label{state_est}}
\begin{figure}[bht]
		\centering
    \includegraphics[width=1.0\columnwidth]{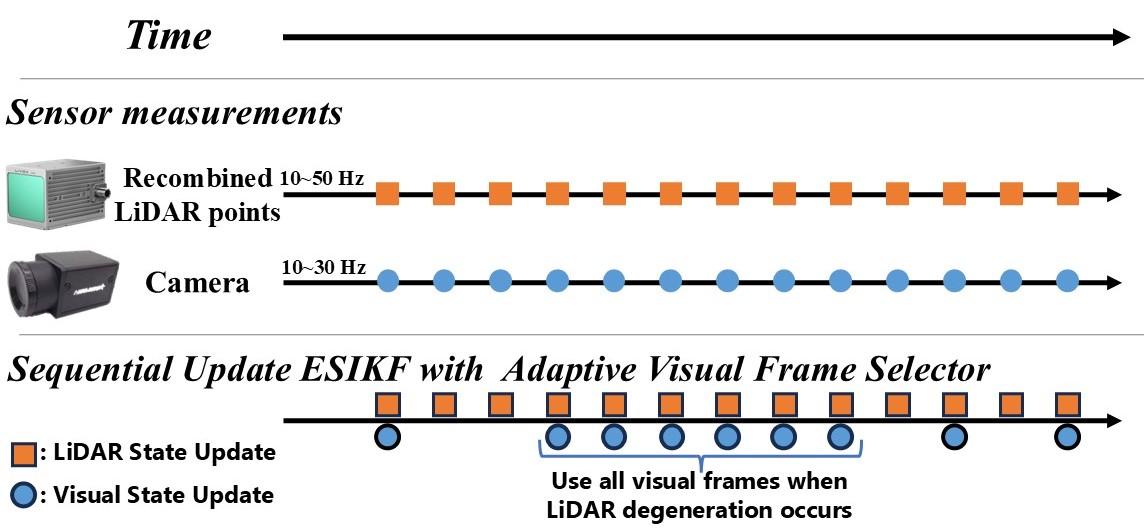}
		\caption{Illustration of sequential update ESIKF with adaptive visual frame selector.}
		\label{esikf_framework}
		\vspace{-0.4cm}
\end{figure}

We first introduce our LiDAR degeneration evaluation method (Section \ref{lidar_degenerate_eva}), followed by the adaptive visual frame selector (Section \ref{ada_img_rate}) enabling more efficient ESIKF-based state estimation, as depicted in Fig.~\ref{esikf_framework}. 

\subsubsection{LiDAR Degeneration Evaluation}\label{lidar_degenerate_eva}
\
\par
After obtaining $N$ undistorted points ${}^L\mathbf{p} \in \mathbf{R}^{3 \times N}$ from one recombined LiDAR scan, where each point is assumed to lie exactly on a plane, the point-to-plane LiDAR measurement model can be expressed as follows:
\begin{equation}\label{eq:lidar_residual}
\mathbf{0} = \left( \mathbf{n}^{gt}\right)^T \left( {}^G\mathbf{T}_L \left( {}^L\mathbf{p} - \delta {}^L\mathbf{p} \right) - \mathbf{q}^{gt}\right),
\end{equation}
where ${}^G\mathbf{T}_L$ denotes the pose of the LiDAR frame with respect to the global frame. The term $\delta {}^L\mathbf{p}$ accounts for the beam noise between the measured ${}^L\mathbf{p}$ and its ground truth. Ideal corresponding planes are parameterized by normal vector $\mathbf{n}^{gt}\in\mathbf{R}^{3 \times N}$ and point on the plane $\mathbf{q}^{gt}\in\mathbf{R}^{3 \times N}$. Additionally, $\mathbf{n}^{est}$ and $\mathbf{q}^{est}$ are the estimated plane features used in the subsequent LiDAR degeneration analysis. 

In the context of LiDAR degeneration evaluation, it is exceedingly rare for all three translational degrees of freedom (DoFs) to remain well-constrained while the three rotational DoFs experience degeneration (e.g., within an ideal spherical structure). Based on this observation, we simplify the problem by focusing solely on the translational constraints.

State estimation is performed by minimizing the residuals from Eq.~\eqref{eq:lidar_residual} as follows:
\vspace{-0.2cm} 
\begin{equation}
\arg\min_{{}^G\mathbf{R}_L,{}^G\mathbf{t}_L} \left\| \underbrace{(\mathbf{n}^{est})^T}_\mathbf{A} \underbrace{{}^G\mathbf{t}_L}_\mathbf{x} + \underbrace{(\mathbf{n}^{est})^T{}^G\mathbf{R}_L{}^L\mathbf{p} - \mathbf{q}}_\mathbf{b} \right\|^2.
\end{equation}
The degeneration of LiDAR can be analyzed by evaluating the stability of this associated optimization problem. The objective function, expressed as \(\left\|\mathbf{A}\mathbf{x} - \mathbf{b} \right\|^2\), can be evaluated the solution stability by analyzing the singular values of \(\mathbf{A}^T\mathbf{A}\), as inspired by \cite{zhang2016degeneracy}. To enhance generalizability across different LiDAR types and parameter settings (e.g., LiDAR scan downsampling strategies), the singular value vector is further normalized as follows:
\vspace{-0.05cm}
\begin{equation}
[\tilde{\sigma}_{\min}, \tilde{\sigma}_{\text{mid}}, \tilde{\sigma}_{\max}] = \text{Normalize}(\text{SVD}( \mathbf{n}^{\text{est}} (\mathbf{n}^{\text{est}})^T)).
\end{equation}
 The smallest normalized value, $\tilde{\sigma}_{\min}$ is compared against a predefined threshold. If $\tilde{\sigma}_{\min}$ remains below this threshold for multiple consecutive frames, the LiDAR module is considered to be in a degenerate state; otherwise, the environment is deemed to provide sufficient constraints for LiDAR scans.

\subsubsection{Adaptive Visual Frame Selector}\label{ada_img_rate}
\
\par
As shown in Fig.~\ref{overview_framework}, we achieve an adaptive image selector based on the degeneration state of the current LiDAR scans and the motion of the sensor. Specifically, when LiDAR degeneration occurs (e.g., when small-FoV AVIA LiDAR faces a wall), all available images are utilized to construct as many constraints as possible, preventing localization failure. Under normal conditions, only sparsely distributed keyframes are selected for updating state and visual related obeservation in map at a reduced frequency, minimizing computational costs.

Keyframes are defined as images captured when the sensor's pose (position or orientation) changes significantly, exceeding a threshold $\boldsymbol{\tau}$ relative to the previous frame. This threshold is adaptively adjusted as:
\begin{equation}
\boldsymbol{\tau} = \sqrt{3} \cdot \tilde{\sigma}_{\min} \cdot \boldsymbol{\tau}_{\text{predefined}},
\end{equation}
where \(\boldsymbol{\tau}_{\text{predefined}} \in \mathbf{R}^{2}_{+}\) contain predefined thresholds for position and orientation. The factor \(\sqrt{3} \cdot \tilde{\sigma}_{\min} \in [0,1]\), scaling the keyframe selection threshold adaptively since \(\tilde{\sigma}_{\min}\) reflects the sufficiency of environmental localization constraints.

\subsection{Map Management\label{map_manager}}

\subsubsection{Map Structure}
The entire map is divided into a unified local map and a long-term visual map. As depicted in the right part of Fig. \ref{overview_framework}, the map, consist of adaptive voxels, is managed by one hash table. Specifically, the Hash value here is a global coordinate index of the root voxel centre \(\mathbf{LOCATION} \in \mathbb{N}^3\) for easier retrieval. The fixed size of root voxel is set to $0.5\times0.5\times0.5$ meters. 


\textbf{Unified Local Map} contains dense points and plane feature (i.e., plane center, normal vector and associated uncertainty) at various scales within a three-level octree structure. Some points selected as visual points are attached with a three-level pyramid of image patches 
for visual state estimation, while the remaining points retain only geometric information for LiDAR scan-to-map matching and subsequent LiDAR state estimation.

\textbf{Long-term Visual Map\label{Visual Global Map Data Structure}} is a collection of historical points with visual observations for visual frame-to-map match. This map is relatively sparse in space, allowing long-term environmental observations to be stored with relatively small memory consumption, while providing essential historical data for long-term localization. 


\begin{figure}[htp]
    \vspace{-0.1cm}
    \centering
    \includegraphics[width=1.0\linewidth]{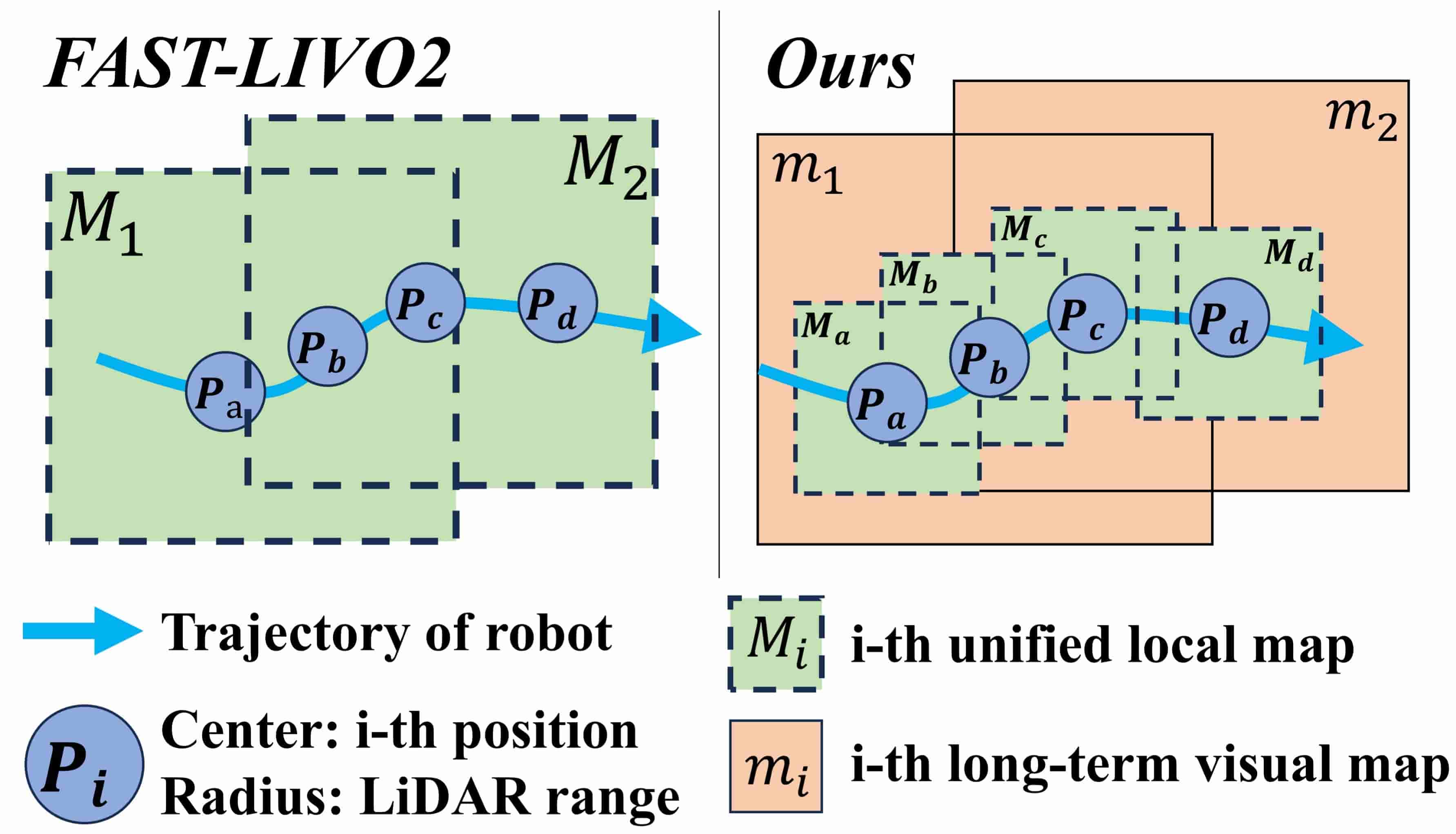}
    \caption{Illustration of map structure and sliding process in our system and FAST-LIVO2.}
    \label{fig:map_sliding}
    \vspace{-0.2cm}
\end{figure}

\subsubsection{Map Updating and Sliding} The process of registering newly observations and estimated plane features to the unified local map follows the approach outlined in \cite{zheng2024fast,yuan2022efficient}. Shown in Fig. \ref{fig:map_sliding}, instead of shifting the large-sized local map only when the sensor range reaches the map boundary, as done in FAST-LIVO2, we adopt a relatively small-sized unified local map with a more frequent map-sliding strategy. Specifically, when the robot's positional movement surpasses a predefined threshold since the last map sliding, plane features and point clouds outside the local map boundary are cleared to maintain efficiency. Points with visual observations that fall outside the small unified local map are transferred to the long-term visual map. The long-term visual map follows a similar sliding mechanism but operates on a larger scale to handle the extended range.
\section{Experiment and Results\label{sec:exp}}
In this section, we first demonstrate the overall performance of our system on public and private datasets, fwollowed by module-specific evaluations.
\subsection{Setup of the Experiments} 


\subsubsection{Public Dataset}
We evaluate our method using the Hilti datasets\cite{zhang2022hilti, helmberger2022hilti}, which feature indoor and outdoor sequences captured in challenging environments such as construction sites, offices, and basements. These datasets include handheld and robot-mounted configurations with LiDAR, cameras, and IMUs operating at different frequencies.


\subsubsection{Private Dataset}
We validated the system's robustness using a B/W camera with a fisheye lens and a Livox Mid-360 LiDAR, synchronized via an STM32 microcontroller. In extremely dark environments, a 15 W onboard illuminator was used to enhance visibility.
\begin{figure}[htb]
    \centering
    \includegraphics[width=0.6\linewidth]{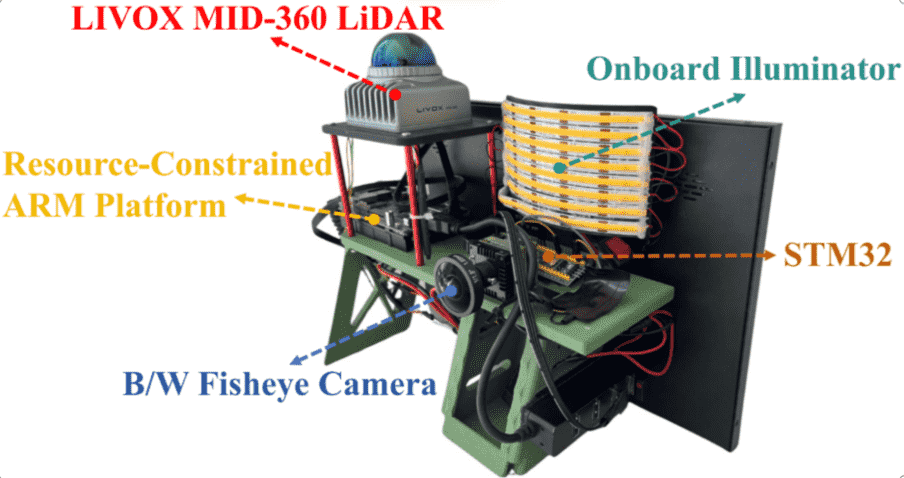}
    \caption{Our platform with hardware synchronization for data acquisition. 
    }
    \label{fig:enter-label}
\end{figure}

\begin{table*}[thbp]
    \vspace{-0.7cm}
    \centering
    \caption{ATE (RMSE, m) for Hilti datasets}
    \begin{threeparttable}
    \fontsize{4}{4}\selectfont 
    \resizebox{1.0\textwidth}{!}{
    \begin{tabular}{clccccccc}
         \hline
         \rule{0pt}{1.0em} \multirow{2}{*}{\textbf{Dataset}} & \multirow{2}{*}{\textbf{Sequence}} & \multirow{2}{*}{\textbf{Ours}}  & \textbf{FAST-}  & \textbf{FAST-} & \textbf{FAST-} & \multirow{2}{*}{\textbf{R3LIVE}} & \textbf{SDV-}  & \textbf{LVI-}  \\
        & &  & \textbf{LIVO2}  & \textbf{LIO2} & \textbf{LIVO} &  & \textbf{LOAM} & \textbf{SAM}  \\
        \hline
        \multirow{8}{*}{\textbf{Hilti '22}} 
        & Construction Ground & \textbf{0.010}  & \textbf{0.010}  & \underline{0.013} & 0.022 & 0.021 & 25.121 & $\times$    \\
        & Construction Multilevel & \underline{0.023}  & \textbf{0.020} & 0.044 & 0.052 & 0.024 & 12.561 & $\times$  \\
        & Construction Stairs & \underline{0.170} & \textbf{0.016}  & 0.320 & 0.241 & 0.784 & 9.212 & 9.142\\
        & Long Corridor & \textbf{0.054} & 0.067 & \underline{0.064} & 0.065 & 0.061 & 19.531 & 6.312  \\
        & Cupola & \underline{0.220}  & \textbf{0.121}  & 0.250 & 0.182 & 2.142 & 9.321 & $\times$ \\
        & Lower Gallery & \underline{0.018}  & \textbf{0.007} &  0.024 & 0.022 & 0.008 & 11.232 & 2.281 \\
        & Attic to Upper Gallery & \underline{0.180}  & \textbf{0.069} & 0.720 & 0.621 & 2.412 & 4.551 &$\times$  \\
        & Outside Building & 0.041  & \underline{0.035} &  \textbf{0.028} & 0.052 & 0.029 & 2.622 & 0.952 \\
        \hline
        \multirow{8}{*}{\textbf{Hilti '23}} 
        & Floor 0 & 0.032  & \textbf{0.021}&  0.031 & \underline{0.022} & 0.024 & 4.621  & $\times$ \\
        & Floor 1 & \textbf{0.018}  & 0.023 &  0.031 &  \underline{0.022} & 0.024 & 7.951 & 8.682 \\
        & Floor 2 & \underline{0.038}  & \textbf{0.022}&  0.083 & 0.048 & 0.046 & 7.912 & $\times$ \\
        & Basement & \underline{0.024} &\textbf{0.016}&  0.038 & 0.035 & \underline{0.024} & 6.151 & $\times$ \\   
        & Stairs & \textbf{0.012} &\underline{0.018} & 0.170 & 0.152 & 0.110 & 9.032 & 3.584\\
        & Parking 3x Floors Down & \underline{0.095}  &\textbf{0.032} & 0.320 & 0.356 & 0.462 & 19.952 & $\times$ \\
        & Large room &  \underline{0.028} &\textbf{0.026} & \underline{0.028} & 0.031 & 0.035 & 16.781 & 0.563 \\
        & Large room (dark) & \underline{0.044}  & 0.046 & \textbf{0.040} & 0.053 & 0.059 & 15.012 & $\times$ \\
        \hline
        \rule{0pt}{1.0em}
        \textbf{Average}
        &  
        & \underline{0.063} 
        & \textbf{0.034} 
        & 0.138 
        & 0.123 
        & 0.391 
        & 11.347
        & 4.502 \\
         \hline
    \end{tabular}
    }
    \begin{tablenotes}
        \footnotesize
         \item[$\times$] denotes the system totally failed.
          \item[\underline{\hspace{0.2cm}}] Underline denotes the second-best, next to the bolded result.
    \end{tablenotes}
    \end{threeparttable}
    \label{rmse_public}
    \vspace{-0.6cm}
\end{table*}

\subsubsection{Experimental Settings and Configurations} The experiments were conducted on two hardware platforms, referred to as \textbf{x86} and \textbf{ARM} platforms for simplicity. The \textbf{x86} platform is a personal laptop with a 13th Gen Intel Core i9-13900HX CPU. The \textbf{ARM} platform, priced at approximately \$100, is notably more cost-effective and features an RK3588 processor with an octa-core architecture (4×Cortex-A76 + 4×Cortex-A55) and a maximum frequency of 2.4GHz.

{A consistent degeneration detection threshold of 0.07 was used across all tests. The local unified map and the long-term visual map were configured with typical edge lengths of 200m and 800m, uniformly across all sequences, with map sliding thresholds of 20m and 100m of robot motion, respectively. Additionally, the keyframe selection thresholds were adapted to the scene scale, with values of 1m-60° for all indoor sequences and 2m-60° for all outdoor sequences. The equidistant projection model was employed to handle fisheye camera images in our private dataset. Other parameters, such as noise settings for different LiDAR models, were inherited from FAST-LIVO2, which had been carefully tuned for robust performance.}

\subsection{System Performance on Public Datasets\label{sec:benchmark}}
\subsubsection{Accuracy on Public Datasets}
In this experiment, we first validated the accuracy of our method on 16 sequences from the Hilti'22 and Hilti'23 datasets, benchmarking it against state-of-the-art systems, including R3LIVE~\cite{lin2022r}, FAST-LIO2~\cite{xu2020fastlio}, SDV-LOAM~\cite{yuan2023sdv}, LVI-SAM~\cite{shan2021lvi}, and our previous work, FAST-LIVO series~\cite{zheng2022fast,zheng2024fast}.
{We made efforts to tune the parameters of all methods to achieve their best performance for comparison.}
Using Hilti official evaluation tools\footnote{https://submit.hilti-challenge.com/}, the RMSE results are shown in Table \ref{rmse_public}. 
From the experimental results, our system achieved the second-highest average RMSE accuracy of 6.3 cm, surpassed only by FAST-LIVO2’s 3.4 cm, while outperforming the state-of-the-art LiDAR-inertial only method FAST-LIO2.
The integration of the visual module within our system proves beneficial, as it enhances localization accuracy. 

Specifically, in sequences with limited visual texture (e.g., \textit{Long Corridor} and \textit{Stairs}), our system outperformed FAST-LIVO2 by selectively reducing visual usage when LiDAR constraints were robust. However, in visually challenging scenarios (e.g., the overexposed \textit{Outside Building} or the poorly lit \textit{Large Room (dark)}), the system exhibited slightly reduced accuracy compared to FAST-LIO2. On the Hilti datasets, SDV-LOAM demonstrated significant limitations, primarily due to the absence of tight IMU integration, which caused notable drift within its LiDAR Odometry subsystem. Additionally, the system's loose coupling between LiDAR and visual measurements, along with insufficient initialization of Visual Odometry, frequently led to convergence issues, such as local optima or failed optimizations. LVI-SAM experienced failures in nine sequences, as its feature-based LIO and VIO subsystems did not fully utilize raw measurements, making it less robust in environments with subtle geometric or texture features (e.g., minimally textured or geometrically simple scenes). While R3LIVE achieved generally strong results, its performance deteriorated in sequences involving intense rotations and sparse structural information (e.g., \textit{Construction Stairs, Cupola} and \textit{Attic to Upper Gallery}). In these scenarios, inadequate pose priors during map alignment caused convergence to suboptimal solutions, ultimately resulting in optimization failures.

\begin{table}[htb]
    \centering
    \vspace{-0.2cm}
    \caption{The mean and standard error of time consumption in Hilti sequences (ms)}
    \resizebox{0.48\textwidth}{!}{%
    \begin{tabular}{ccccc}
     \hline
     \textbf{Architecture} &  \multicolumn{2}{c}{\textbf{ARM}} & \multicolumn{2}{c}{\textbf{x86 PC}} \\
     \cmidrule(lr){1-5}
     \textbf{SLAM System} & \textbf{Ours} & \textbf{FAST-LIVO2} & \textbf{Ours} & \textbf{FAST-LIVO2} \\
      \hline
     \textbf{LiDAR Part} & 53.83 / 5.35 &  56.50 / 5.56  & 23.36 / 2.36 & 25.05 / 2.41 \\
     \textbf{Visual Part} & 3.99 / 1.80 & 19.36 / 4.75 & 2.64 / 1.12 & 10.61 / 2.68 \\
      \hline
     \rule{0pt}{1.0em} \textbf{Total} & 57.82 / 6.75 & 75.87 / 9.77 & 25.99 / 3.22 & 35.66 / 4.88 \\
      \hline
    \end{tabular}%
    }
    \label{tab:time_consumption}
    \vspace{-0.3cm}
\end{table}

\begin{figure*}[t]
    \centering
    \includegraphics[width=1.0\linewidth]{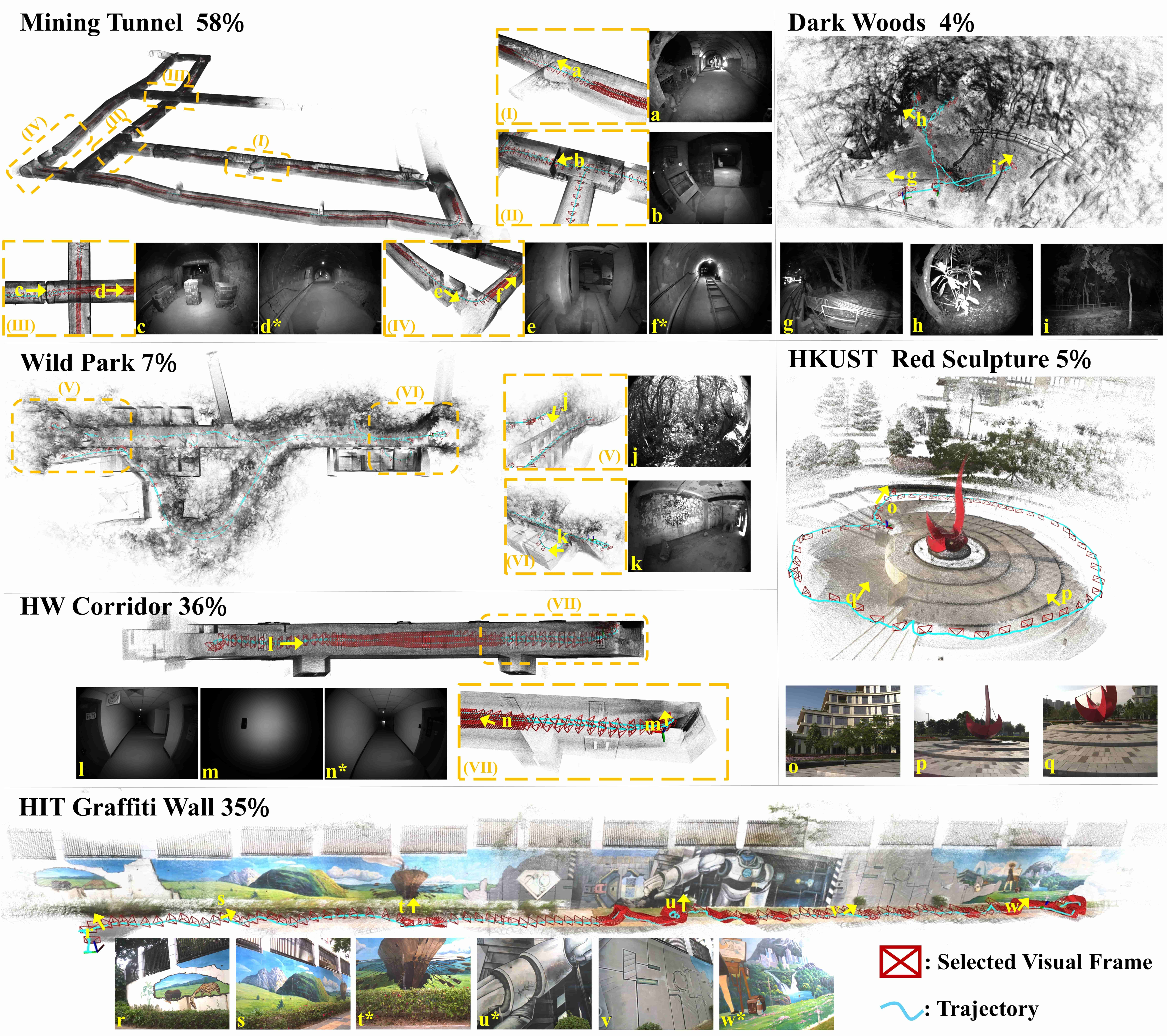}
    \caption{Visualization of some representative tested sequences. * denotes the moment when detected LiDAR degeneration occurs. The percentage next to each sequence name indicates the proportion of selected images relative to all available images in that sequence.}
    \label{fig:fig_seqences}
    \vspace{-0.7cm}
\end{figure*}

\subsubsection{Computation Efficiency on Public Datasets} 
We evaluated the algorithm's runtime on 16 sequences from the Hilti dataset using both x86 and ARM platforms\footnote{Runtime on the ARM platform was measured entirely on the CPU without hardware acceleration.}, as summarized in Table~\ref{tab:time_consumption}. Generally, our system achieved a notable improved computational efficiency compared to FAST-LIVO2. This improvement is primarily attributed to the adaptive frame rate strategy in the visual module, which significantly reduced the runtime of visual processing. Additionally, the LiDAR module experienced a slight increase in runtime due to optimizations in the code implementation.


\begin{table}[htbp]
\centering
\caption{Memory usage comparison between our algorithm and FAST-LIVO2 for Hilti dataset sequences (gb)}
\label{tab:mem_usage}
\fontsize{8}{8}\selectfont
\begin{tabular}{clcc}
\hline
\rule{0pt}{1.0em} \textbf{Dataset} & \textbf{Sequence} & \textbf{Ours} & \textbf{FAST-LIVO2} \\
\hline
\multirow{8}{*}{\textbf{Hilti '22}}
& Construction Ground         & 3.0 & 4.1 \\
& Construction Multilevel     & 3.7 & 4.9 \\
& Construction Stairs         & 1.3 & 2.4 \\
& Long Corridor               & 0.8 & 1.2 \\
& Cupola                      & 1.9 & 3.4 \\
& Lower Gallery               & 1.0 & 1.6 \\
& Attic to Upper Gallery      & 1.4 & 2.4 \\
& Outside Building            & 2.4 & 3.0 \\
 \hline
\multirow{8}{*}{\textbf{Hilti '23}}
& Floor 0                     & 2.4 & 3.1 \\
& Floor 1                     & 1.5 & 2.2 \\
& Floor 2                     & 1.7 & 2.3 \\
& Basement                    & 1.3 & 2.4 \\
& Stairs                      & 1.3 & 1.8 \\
& Parking 3x Floors Down      & 1.7 & 2.6 \\
& Large room                  & 1.0 & 1.6 \\
& Large room (dark)           & 1.1 & 1.6 \\
 \hline
\rule{0pt}{1.0em} \textbf{Average} &            & 1.7 & 2.5 \\
 \hline
\end{tabular}
\vspace{-0.4cm}
\end{table}

\subsubsection{Memory Consumption on Public Datasets}We further evaluate our method's computational efficiency and memory consumption on the public dataset. Since FAST-LIVO2 demonstrated the highest accuracy among all compared methods, as shown in Table~\ref{rmse_public}, we focus on a comparison between it with our proposed system to highlight the improvements in resource efficiency. We measure the memory usage of the validated Hilti sequences on x86 platform, summarized in the table~\ref{tab:mem_usage}. It can be observed that our system significantly reduces memory consumption compared to FAST-LIVO2. Since the algorithm's memory consumption is primarily attributed to map storage, the reduction in small-scale scenarios is largely due to the adaptive image frame selector, which effectively minimizes redundant visual observations in the map. For larger-scale scenarios, the memory reduction is primarily attributed to our designed map structure, which utilizes a smaller-sized local unified map to efficiently manage storage.
\subsection{System Performance on Private Datasets}
We conducted qualitative tests on the system's robustness and computational real-time performance using private datasets.
\subsubsection{Experiments on Challenging Private Sequences} We conducted tests on challenging private sequences. Illustrated in Fig.~\ref{fig:fig_seqences}, these sequences encompass a variety of SLAM challenges, such as dynamic lighting conditions in \textit{Mining Tunnel}, extremely low illumination in \textit{Dark Woods}, unstructured environments in \textit{Wild Park} and severe LiDAR degeneration in \textit{HIT Graffiti Wall}. Additionally, they cover both indoor and outdoor scenes captured at different times, also including indoor scenarios with lights turned off (e.g., \textit{HW Corridor}). As an experimental result, our system demonstrated robust performance, producing sharp point cloud maps and even vivid colourful point clouds. In all sequences where the data collection equipment physically returns to the starting position, our system achieves low return-to-origin drift of less than 2 cm.

Meanwhile, the percentage in Fig.~\ref{fig:fig_seqences} refers to the proportion of selected images relative to all available images, showing the system's adaptive sensitivity to the environmental constraints provided by LiDAR. For instance, in long tunnel-like corridors such as the \textit{Mining Tunnel} and mid part of \textit{HW Corridor}, dense visual frames are selected due to weak LiDAR constraints. In contrast, in environments with strong LiDAR constraints, such as the \textit{HKUST Red Sculpture}, images are selected more sparsely in space.

\subsubsection{Real-Time Localization Experiments on Resource-Constrained ARM Platform}
\begin{figure}
    \centering
    \includegraphics[width=1.0\linewidth]{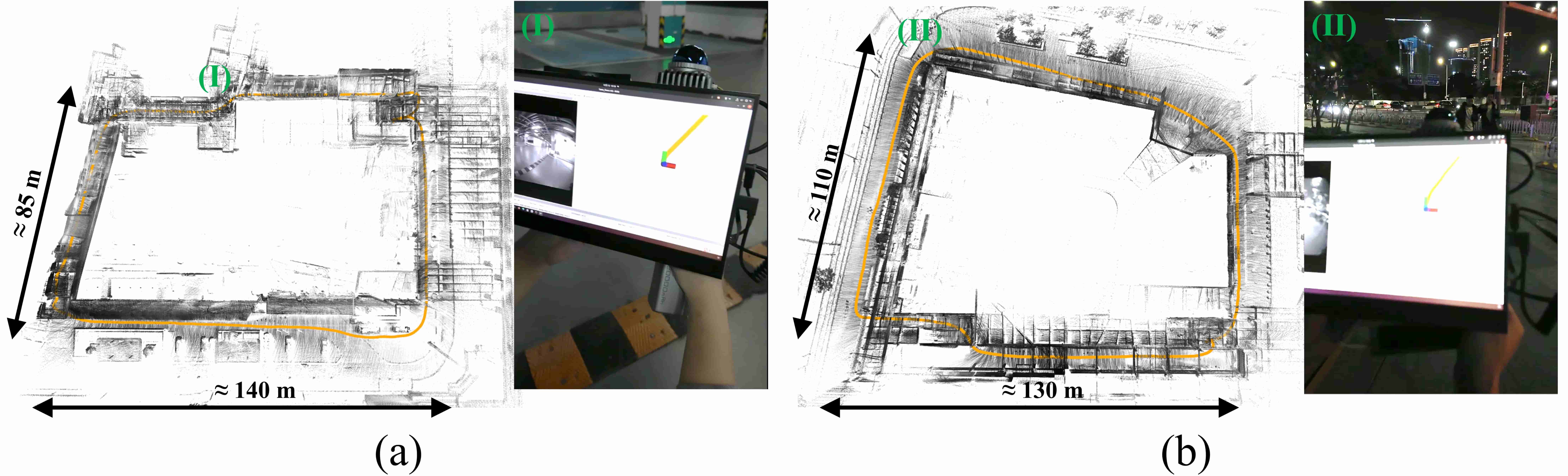}
    \caption{Real-time localization experiments on the RK3588 platform. (a) Underground parking lot and (b) nighttime street. For each scenario, the left image shows the trajectory and point cloud, and the right image shows a third-person view at a specific moment.}
    \label{fig:arm_exp}
    \vspace{-0.6cm}
\end{figure}

Shown in Fig.~\ref{fig:arm_exp}, we conducted real-time localization tests in the evening on bustling streets with pedestrian and in an underground parking lot with vehicles exiting, as well as substantial indoor-outdoor lighting variations. All computations were performed entirely on the onboard ARM platform. The results demonstrated robust localization with a per-frame time cost of 37 ms.
\begin{figure}[htb]
    \centering    
    \includegraphics[width=1.0\linewidth]{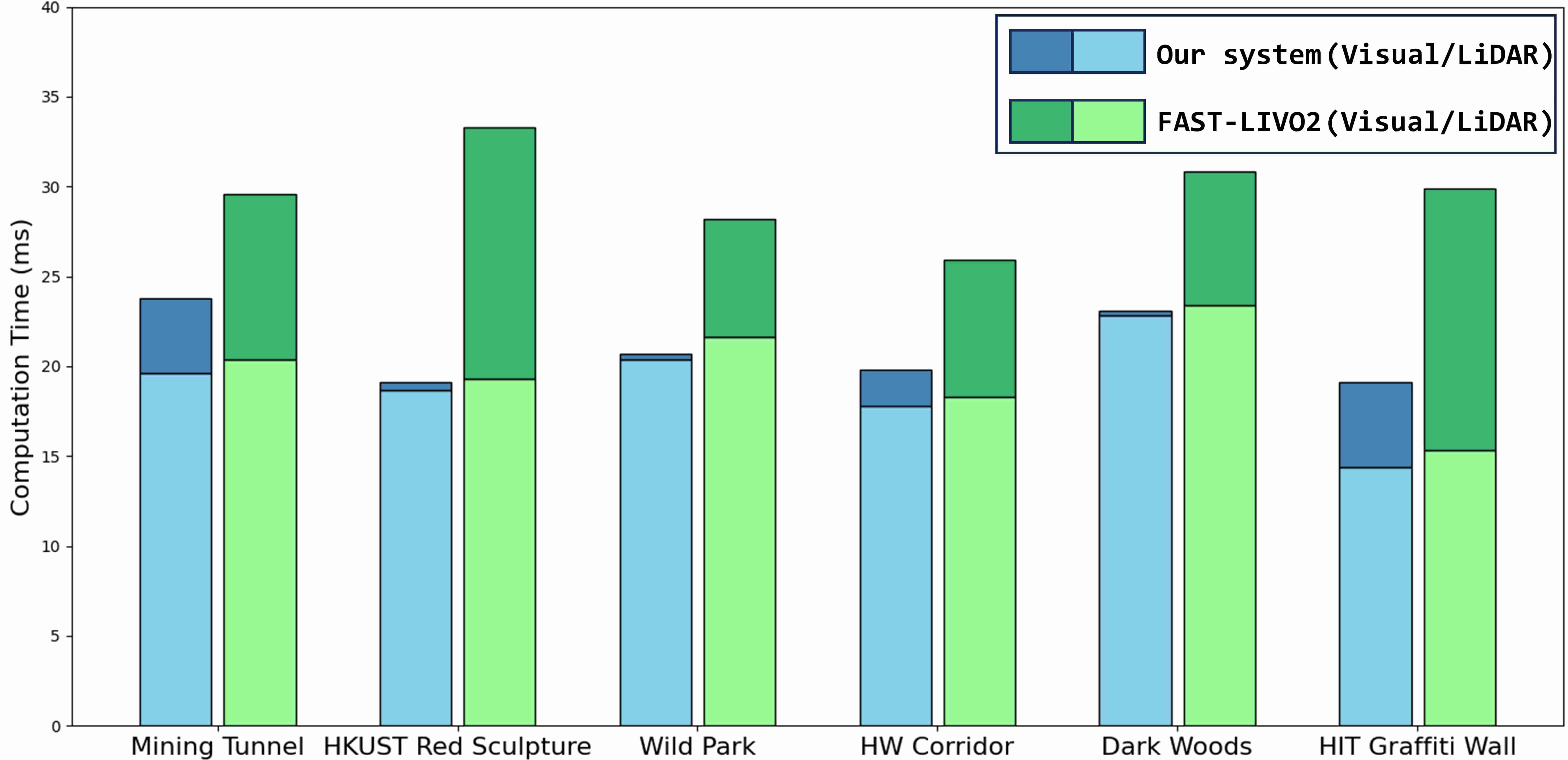}
    \caption{LiDAR and visual processing times for our method and FAST-LIVO2 in seven scenarios. LiDAR time forms the base, with visual time stacked above.}
    \label{fig:compute_time}
    \vspace{-0.6cm}
\end{figure}
\subsection{Evaluation of Adaptive Image Frame Selector\label{sec:robust_exp}}
\subsubsection{Evaluation of Computational Efficiency Optimization via Adaptive Image Frame Selector}
From the visualization results of the selected visual frames in Fig. \ref{fig:fig_seqences}, it can be observed that different proportions of images were selected from all available visual measurements across various sequences, based on the LiDAR degeneration conditions in the environment. We conducted a detailed analysis of the computation times for the LiDAR and visual modules in these sequences and compared them with FAST-LIVO2. As shown in the Fig. \ref{fig:compute_time}, the computation time for the visual module is significantly reduced in all cases. Notably, in sequences with strong LiDAR constraints, the computational overhead of the visual module is exceptionally low.

\subsubsection{Evaluation of LiDAR Degeneration Detection Module}
\begin{figure}[htb]
    \centering
    \includegraphics[width=0.9\linewidth]{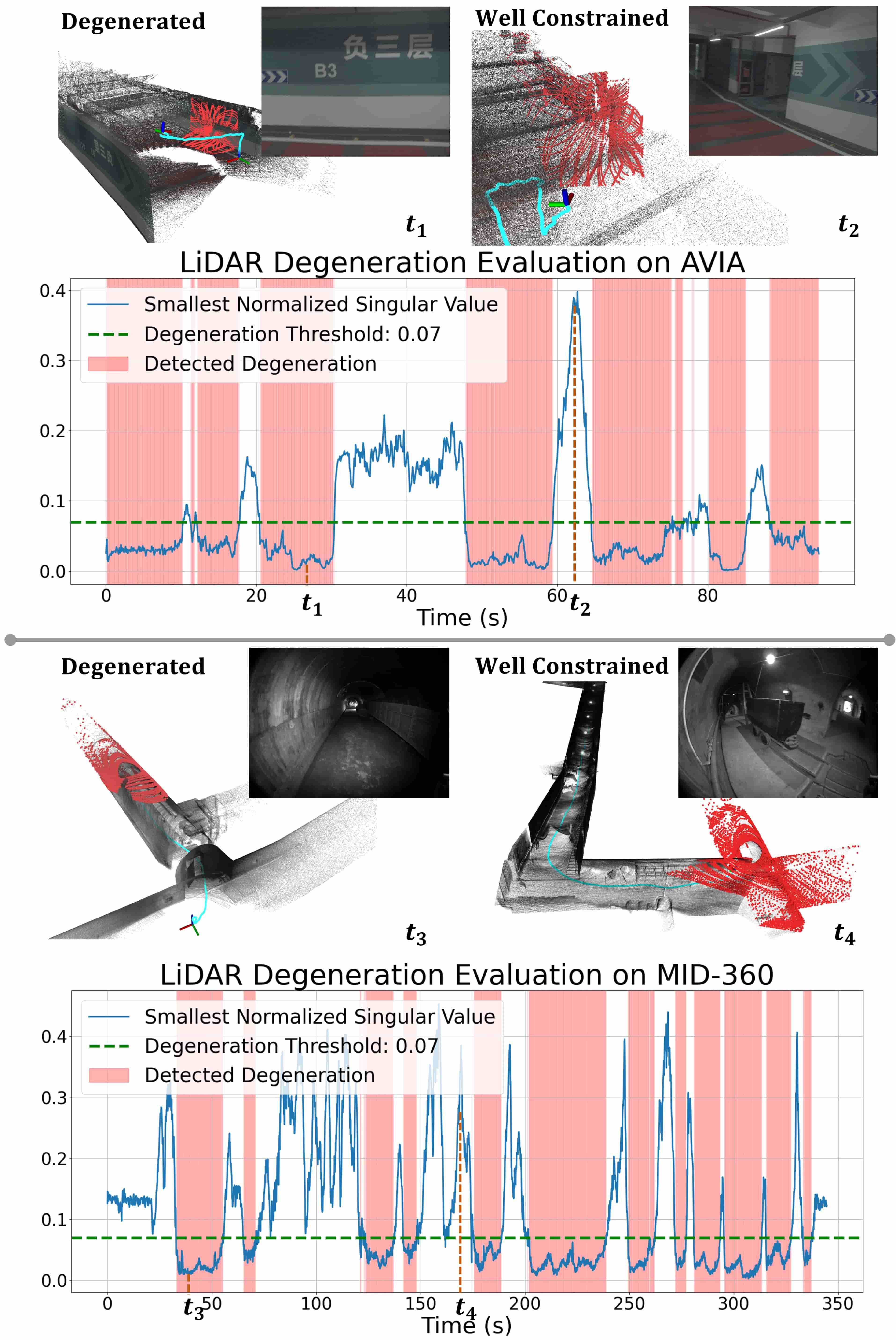}
    \caption{Validation of LiDAR degeneracy evaluation. The plot shows the smallest normalized singular value, with degeneration detected when the value falls below the threshold (dashed line), indicated by red shaded areas. The top visualizations correspond to moments of low ($t_1, t_3$) and high (\(t_2, t_4\)) singular values, with red points representing the current LiDAR scan and the top-right image showing the corresponding first-person camera view.}
    \label{lidar_dege1}
    \vspace{-0.3cm}
\end{figure}

As illustrated in Fig.~\ref{lidar_dege1}, we conducted a detailed analysis of the effectiveness of the LiDAR degeneration detection module using two highly challenging sequences for LiDAR. The results highlight the algorithm's sensitivity in identifying degeneration, as evidenced by extremely low eigenvalues in degenerate scenarios (e.g., the Aivia LiDAR facing a wall at $t_1$ and the Mid360 LiDAR in a long, straight tunnel at $t_3$). Furthermore, the analysis confirms that a universal degeneration threshold can be applied across different LiDAR types, demonstrating the effectiveness of our LiDAR degeneration evaluation module.

\vspace{-0.2cm}

\subsection{Evaluation of Map Structure}
To evaluate the proposed long-term visual map, we perform an ablation study to assess its impact on localization accuracy and memory usage using the large-scale sequence \textit{HKIsland03} in MARS-LVIG datast\cite{li2024mars}.           
\begin{figure}[htb]
    \centering
    \includegraphics[width=1.0\linewidth]{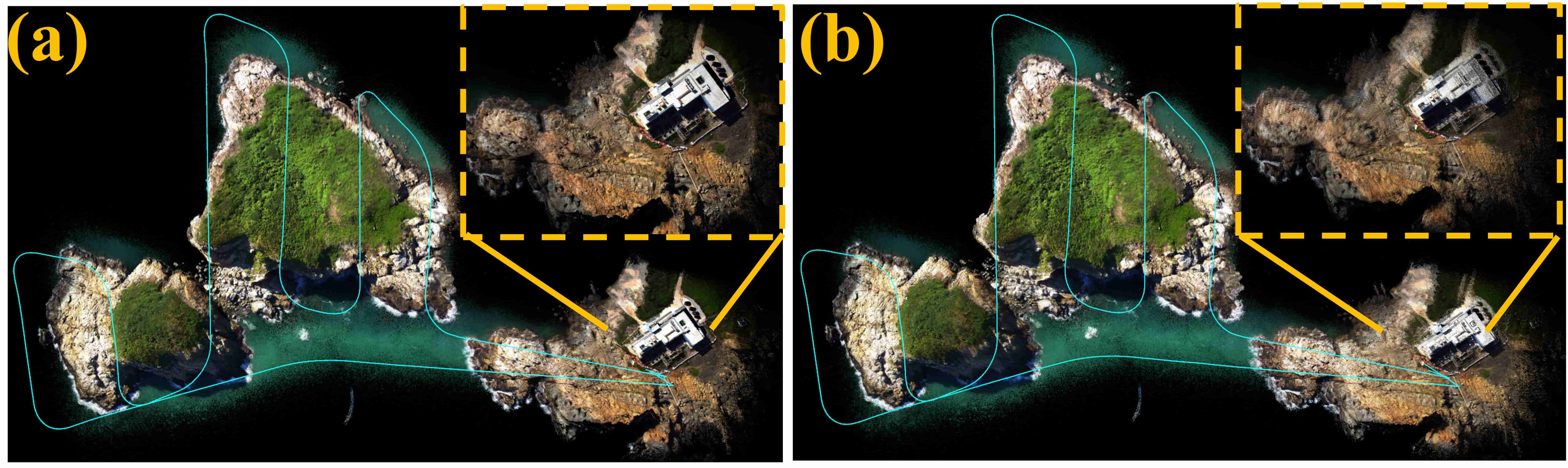}
    \caption{Ablation study results on our proposed long-term visual map structure, presented as reconstructed colored point clouds. (a) With the proposed long-term visual map. (b) Without the long-term visual map. }
    \label{fig:Ablation}
    \vspace{-0.3cm}
\end{figure}
\subsubsection{Accuracy Comparison}
The localization accuracy is evaluated using the RTK trajectory as ground truth. We evaluated RMSE using the evo toolkit\footnote{https://github.com/MichaelGrupp/evo}. In this sequence, the long-term visual map improved accuracy, achieving better RMSE 0.39m related to 0.85m, as evidenced by the detailed point cloud visualization in Fig.~\ref{fig:Ablation}.
\begin{figure}[htp]
    \centering
    \includegraphics[width=0.8\linewidth]{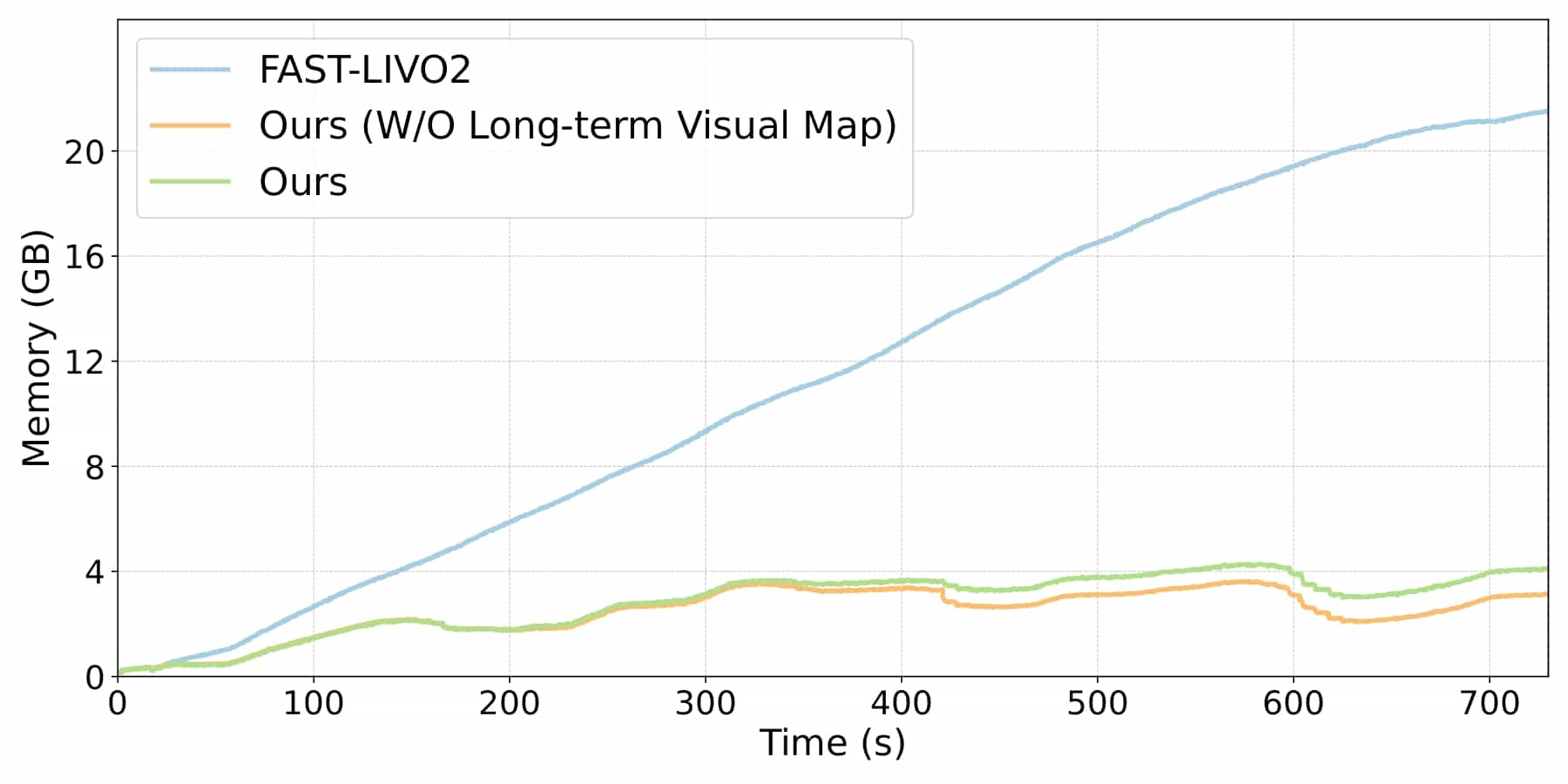}
    \caption{Comparison of memory usage for different algorithms in MaRS LVIG sequence \textit{HKIsland03}. }
    \label{fig:mem_usage}
    \vspace{-0.3cm}
\end{figure}
\subsubsection{Memory Consumption Comparison}
Shown in Fig.~\ref{fig:mem_usage}, while the long-term visual map introduces additional memory overhead, its impact on memory usage is relatively modest compared to the significant improvement in localization accuracy. This trade-off is justified, as the enhanced robustness and accuracy, confirming the map’s suitability for large-scale deployments.
{Meanwhile, the unified local map edge length parameter of FAST-LIVO2 was set to 2km, as in the original paper. With the scenario contained within the map boundaries, memory usage continuously increases.}

\vspace{-0.15cm}
\section{Conclusion}
\vspace{-0.1cm}
This paper presents a lightweight LIVO system optimized for resource-constrained platforms. By integrating a degeneration-aware adaptive visual frame selector and a memory-efficient mapping structure, the system significantly reduces computation and memory usage while maintaining odometry accuracy. Extensive experiments validate its robust performance and demonstrate its capability to run efficiently on resource-constrained edge devices, enabling real-time operation in challenging environments.
\vspace{-0.2cm}
\bibliographystyle{IEEEtran}
\bibliography{paper}

\begin{thebibliography}{10}
\providecommand{\url}[1]{#1}
\csname url@samestyle\endcsname
\providecommand{\newblock}{\relax}
\providecommand{\bibinfo}[2]{#2}
\providecommand{\BIBentrySTDinterwordspacing}{\spaceskip=0pt\relax}
\providecommand{\BIBentryALTinterwordstretchfactor}{4}
\providecommand{\BIBentryALTinterwordspacing}{\spaceskip=\fontdimen2\font plus
\BIBentryALTinterwordstretchfactor\fontdimen3\font minus \fontdimen4\font\relax}
\providecommand{\BIBforeignlanguage}[2]{{%
\expandafter\ifx\csname l@#1\endcsname\relax
\typeout{** WARNING: IEEEtran.bst: No hyphenation pattern has been}%
\typeout{** loaded for the language `#1'. Using the pattern for}%
\typeout{** the default language instead.}%
\else
\language=\csname l@#1\endcsname
\fi
#2}}
\providecommand{\BIBdecl}{\relax}
\BIBdecl

\bibitem{zhang2014loam}
J.~Zhang and S.~Singh, ``Loam: Lidar odometry and mapping in real-time.'' in \emph{Robotics: Science and Systems}, vol.~2, no.~9, 2014.

\bibitem{qin2018vins}
T.~Qin, P.~Li, and S.~Shen, ``Vins-mono: A robust and versatile monocular visual-inertial state estimator,'' \emph{IEEE Transactions on Robotics}, vol.~34, no.~4, pp. 1004--1020, 2018.

\bibitem{xu2020fastlio}
W.~{Xu} and F.~{Zhang}, ``Fast-lio: A fast, robust lidar-inertial odometry package by tightly-coupled iterated kalman filter,'' \emph{IEEE Robotics and Automation Letters}, pp. 1--1, 2021.

\bibitem{shan2021lvi}
T.~Shan, B.~Englot, C.~Ratti, and D.~Rus, ``Lvi-sam: Tightly-coupled lidar-visual-inertial odometry via smoothing and mapping,'' in \emph{2021 IEEE international conference on robotics and automation (ICRA)}.\hskip 1em plus 0.5em minus 0.4em\relax IEEE, 2021, pp. 5692--5698.

\bibitem{zheng2022fast}
C.~Zheng, Q.~Zhu, W.~Xu, X.~Liu, Q.~Guo, and F.~Zhang, ``Fast-livo: Fast and tightly-coupled sparse-direct lidar-inertial-visual odometry,'' in \emph{2022 IEEE/RSJ international conference on intelligent robots and systems (IROS)}.\hskip 1em plus 0.5em minus 0.4em\relax IEEE, 2022, pp. 4003--4009.

\bibitem{zheng2024fast}
C.~Zheng, W.~Xu, Z.~Zou, T.~Hua, C.~Yuan, D.~He, B.~Zhou, Z.~Liu, J.~Lin, F.~Zhu \emph{et~al.}, ``Fast-livo2: Fast, direct lidar-inertial-visual odometry,'' \emph{arXiv preprint arXiv:2408.14035}, 2024.

\bibitem{lin2021r2live}
J.~Lin, C.~Zheng, W.~Xu, and F.~Zhang, ``R$^2$live: A robust, real-time, lidar-inertial-visual tightly-coupled state estimator and mapping,'' \emph{IEEE Robotics and Automation Letters}, vol.~6, no.~4, pp. 7469--7476, 2021.

\bibitem{lin2022r}
J.~Lin and F.~Zhang, ``R 3 live: A robust, real-time, rgb-colored, lidar-inertial-visual tightly-coupled state estimation and mapping package,'' in \emph{2022 International Conference on Robotics and Automation (ICRA)}.\hskip 1em plus 0.5em minus 0.4em\relax IEEE, 2022, pp. 10\,672--10\,678.

\bibitem{forster2016svo}
C.~Forster, Z.~Zhang, M.~Gassner, M.~Werlberger, and D.~Scaramuzza, ``Svo: Semidirect visual odometry for monocular and multicamera systems,'' \emph{IEEE Transactions on Robotics}, vol.~33, no.~2, pp. 249--265, 2016.

\bibitem{geneva2020openvins}
P.~Geneva, K.~Eckenhoff, W.~Lee, Y.~Yang, and G.~Huang, ``Openvins: A research platform for visual-inertial estimation,'' in \emph{2020 IEEE International Conference on Robotics and Automation (ICRA)}.\hskip 1em plus 0.5em minus 0.4em\relax IEEE, 2020, pp. 4666--4672.

\bibitem{zhou2020ego}
X.~Zhou, Z.~Wang, H.~Ye, C.~Xu, and F.~Gao, ``Ego-planner: An esdf-free gradient-based local planner for quadrotors,'' \emph{IEEE Robotics and Automation Letters}, vol.~6, no.~2, pp. 478--485, 2020.

\bibitem{lu2022manifold}
G.~Lu, W.~Xu, and F.~Zhang, ``On-manifold model predictive control for trajectory tracking on robotic systems,'' \emph{IEEE Transactions on Industrial Electronics}, vol.~70, no.~9, pp. 9192--9202, 2022.

\bibitem{zhang2016degeneracy}
J.~Zhang, M.~Kaess, and S.~Singh, ``On degeneracy of optimization-based state estimation problems,'' in \emph{2016 IEEE international conference on robotics and automation (ICRA)}.\hskip 1em plus 0.5em minus 0.4em\relax IEEE, 2016, pp. 809--816.

\bibitem{tuna2023x}
T.~Tuna, J.~Nubert, Y.~Nava, S.~Khattak, and M.~Hutter, ``X-icp: Localizability-aware lidar registration for robust localization in extreme environments,'' \emph{IEEE Transactions on Robotics}, 2023.

\bibitem{lim2023adalio}
H.~Lim, D.~Kim, B.~Kim, and H.~Myung, ``Adalio: Robust adaptive lidar-inertial odometry in degenerate indoor environments,'' in \emph{2023 20th International Conference on Ubiquitous Robots (UR)}.\hskip 1em plus 0.5em minus 0.4em\relax IEEE, 2023, pp. 48--53.

\bibitem{zhang2024lio}
T.~Zhang, X.~Zhang, Z.~Liao, X.~Xia, and Y.~Li, ``As-lio: Spatial overlap guided adaptive sliding window lidar-inertial odometry for aggressive fov variation,'' \emph{arXiv preprint arXiv:2408.11426}, 2024.

\bibitem{zhu2023swarm}
F.~Zhu, Y.~Ren, F.~Kong, H.~Wu, S.~Liang, N.~Chen, W.~Xu, and F.~Zhang, ``Swarm-lio: Decentralized swarm lidar-inertial odometry,'' in \emph{2023 IEEE international conference on robotics and automation (ICRA)}.\hskip 1em plus 0.5em minus 0.4em\relax IEEE, 2023, pp. 3254--3260.

\bibitem{zhu2024swarm}
F.~Zhu, Y.~Ren, L.~Yin, F.~Kong, Q.~Liu, R.~Xue, W.~Liu, Y.~Cai, G.~Lu, H.~Li \emph{et~al.}, ``Swarm-lio2: Decentralized, efficient lidar-inertial odometry for uav swarms,'' \emph{IEEE Transactions on Robotics}, 2024.

\bibitem{lee2024switch}
J.~Lee, R.~Komatsu, M.~Shinozaki, T.~Kitajima, H.~Asama, Q.~An, and A.~Yamashita, ``Switch-slam: Switching-based lidar-inertial-visual slam for degenerate environments,'' \emph{IEEE Robotics and Automation Letters}, 2024.

\bibitem{yuan2023sdv}
Z.~Yuan, Q.~Wang, K.~Cheng, T.~Hao, and X.~Yang, ``Sdv-loam: semi-direct visual--lidar odometry and mapping,'' \emph{IEEE Transactions on Pattern Analysis and Machine Intelligence}, vol.~45, no.~9, pp. 11\,203--11\,220, 2023.

\bibitem{wang2021dv}
W.~Wang, J.~Liu, C.~Wang, B.~Luo, and C.~Zhang, ``Dv-loam: Direct visual lidar odometry and mapping,'' \emph{Remote Sensing}, vol.~13, no.~16, p. 3340, 2021.

\bibitem{shi2024point}
T.~Shi, K.~Qian, Y.~Fang, Y.~Zhang, and H.~Yu, ``Point-line livo using patch-based gradient optimization for degenerate scenes,'' \emph{IEEE Robotics and Automation Letters}, 2024.

\bibitem{zhu2021camvox}
Y.~Zhu, C.~Zheng, C.~Yuan, X.~Huang, and X.~Hong, ``Camvox: A low-cost and accurate lidar-assisted visual slam system,'' in \emph{2021 IEEE International Conference on Robotics and Automation (ICRA)}.\hskip 1em plus 0.5em minus 0.4em\relax IEEE, 2021, pp. 5049--5055.

\bibitem{yuan2024sr}
Z.~Yuan, J.~Deng, R.~Ming, F.~Lang, and X.~Yang, ``Sr-livo: Lidar-inertial-visual odometry and mapping with sweep reconstruction,'' \emph{IEEE Robotics and Automation Letters}, 2024.

\bibitem{cai2021ikd}
Y.~Cai, W.~Xu, and F.~Zhang, ``ikd-tree: An incremental kd tree for robotic applications,'' \emph{arXiv preprint arXiv:2102.10808}, 2021.

\bibitem{yuan2022efficient}
C.~Yuan, W.~Xu, X.~Liu, X.~Hong, and F.~Zhang, ``Efficient and probabilistic adaptive voxel mapping for accurate online lidar odometry,'' \emph{IEEE Robotics and Automation Letters}, vol.~7, no.~3, pp. 8518--8525, 2022.

\bibitem{zhang2022hilti}
L.~Zhang, M.~Helmberger, L.~F.~T. Fu, D.~Wisth, M.~Camurri, D.~Scaramuzza, and M.~Fallon, ``Hilti-oxford dataset: A millimeter-accurate benchmark for simultaneous localization and mapping,'' \emph{IEEE Robotics and Automation Letters}, vol.~8, no.~1, pp. 408--415, 2022.

\bibitem{helmberger2022hilti}
M.~Helmberger, K.~Morin, B.~Berner, N.~Kumar, G.~Cioffi, and D.~Scaramuzza, ``The hilti slam challenge dataset,'' \emph{IEEE Robotics and Automation Letters}, vol.~7, no.~3, pp. 7518--7525, 2022.

\bibitem{li2024mars}
H.~Li, Y.~Zou, N.~Chen, J.~Lin, X.~Liu, W.~Xu, C.~Zheng, R.~Li, D.~He, F.~Kong \emph{et~al.}, ``Mars-lvig dataset: A multi-sensor aerial robots slam dataset for lidar-visual-inertial-gnss fusion,'' \emph{The International Journal of Robotics Research}, p. 02783649241227968, 2024.

\end{thebibliography}

\end{document}